\documentclass{llncs}
\usepackage[utf8]{inputenc}

\usepackage{amsmath}
\usepackage{amssymb}
\usepackage{tikz}
\usetikzlibrary{bayesnet}

\newcommand*{\cequal}{\overset{c}{=}}

\title{Diffeomorphic brain shape modelling \newline using Gauss-Newton optimisation}
\author{Yaël Balbastre \and Mikael Brudfors \and Kevin Bronik \and John Ashburner}
\institute{Wellcome Centre for Human Neuroimaging, University College London, London, UK}

\begin{document}

\setlength{\abovedisplayskip}{3pt}
\setlength{\belowdisplayskip}{3pt}

\maketitle

\begin{abstract}
    Shape modelling describes methods aimed at capturing the natural variability of shapes and commonly relies on probabilistic interpretations of dimensionality reduction techniques such as principal component analysis. Due to their computational complexity when dealing with dense deformation models such as diffeomorphisms, previous attempts have focused on explicitly reducing their dimension, diminishing \emph{de facto} their flexibility and ability to model complex shapes such as brains. In this paper, we present a generative model of shape that allows the covariance structure of deformations to be captured without squashing their domain, resulting in better normalisation. An efficient inference scheme based on Gauss-Newton optimisation is used, which enables processing of 3D neuroimaging data. We trained this algorithm on segmented brains from the OASIS database, generating physiologically meaningful deformation trajectories. To prove the model's robustness, we applied it to unseen data, which resulted in equivalent fitting scores.
\end{abstract}

\section{Introduction}

In neuroimaging studies, or more generally in shape analysis, normalising a set of subjects consists in deforming them towards a common space that allows one-to-one correspondence. Finding this common space usually reduces to finding an optimal shape in terms of distance to all subjects in the space of deformations. However, the covariance structure of these deformations is not known \emph{a priori} and the deformation metric generally involves penalising roughness. Yet, in a Bayesian setting, a prior that is informative of population variance would make the registration process, which relies on \emph{a posteriori} estimates, more robust. Shape models aim to learn this covariance structure from the data. As deformations are parameterised in a very high-dimensional space, learning their covariance is computationally intractable. Dimensionality reduction techniques are therefore commonly used, even though some have tackled this problem by parameterising deformations using location-adaptive control points \cite{Durrleman2013}.

For a long time, due to their computational complexity, shape modelling approaches had only been applied to simple models of deformations \cite{Cootes1994} or simple data sets \cite{Cootes2008,Fletcher2004}. Recently, Zhang and Fletcher applied a probabilistic shape model, named principal geodesic analysis, to densely sampled diffeomorphisms and 3D MR images of the brain \cite{Zhang2015b}. Diffeomorphisms correspond to a particular family of deformations that are ensured to be invertible, allowing for very large deformations. Geodesic shooting of diffeomorphisms involves specifying a Riemannian metric on their tangent space and allows a diffeomorphism to to be entirely parameterised by its initial velocity \cite{Miller2006,Ashburner2011a}. However, because Zhang and Fletcher's optimisation scheme relies either on Gradient descent or on Monte Carlo sampling of the posterior, they have focused on effectively reducing the dimensionality of velocity fields. It results in an effective approach for studying the principle modes of variations, but may give less accurate alignment than with classical approaches. In particular, they do not explicitly model ``anatomical noise'', \emph{i.e.}, deformations that are not captured by the principal modes.

Here, we propose a generative shape model, whose posterior is inferred using variational inference and Laplace approximations. A residual velocity field capturing anatomical noise is explicitly defined and its magnitude is inferred from the data. An efficient Gauss-Newton optimisation is used to obtain the maximum \emph{a posteriori} latent subspace as well as individual coordinates, minimising the chances of falling into local minima, and making the registration more robust.

\section{Methods}

\subsection{Generative shape model}

First, let us define a generative model of brain shape. Here, the observed variables are supposed to be categorical images (\emph{i.e.}, segmentations) comprising $K$ classes --- \emph{e.g.} grey matter, white matter, background --- stemming from a categorical distribution. This kind of data term has proved very effective for driving registration \cite{Ashburner2009} and is compatible with unified models of registration and segmentation. If needed, it is straightforward to replace this term with a stationary Gaussian noise model. The template $\vec{\mu}$ encodes prior probabilities of finding each of the $K$ classes in a given location, and is deformed towards the $n$-th subject according to the spatial transform $\vec{\phi}_n$. In practice, its log-representation $\vec{a}$ is encoded by trilinear basis functions, and the deformed template is recovered by softmax interpolation \cite{Ashburner2009}:
\begin{equation}
    \textstyle
    \small
    \mu_n^{(k)}(\vec{x})
    = \left[\mathrm{softmax}\left(\vec{a}\circ\vec{\phi}_n\left(\vec{x}\right)\right)\right]_k
    = \frac{\exp\left( a^{(k)}\circ\vec{\phi}_n(\vec{x})\right)}{\sum_{l=1}^K \exp\left( a^{(l)}\circ\vec{\phi}_n(\vec{x})\right)} \enspace.
\end{equation}%
Note that the discrete operation $\vec{a}\circ\vec{\phi}_n$ can be equivalently written as the matrix multiplication, $\vec{\Phi}_n\vec{a}$, where $\vec{\Phi}_n$ is a large and sparse matrix that depends on $\vec{\phi}_n$ and performs the combined ``sample and interpolate'' operation. We will name this operation \emph{pulling}, while the multiplication by its transpose, $\vec{\Phi}_n^{T}$, will be named \emph{pushing}.

Let $\left\{ \vec{f}_n \in \mathbb{R}^{I \times K} ~;~ 1 \leqslant n \leqslant N\right\}$ be the set of observed images\footnote{We assume that they all have the same lattice, but this condition can be waived by composing each diffeomorphic transform with a fixed ``change of lattice'' transform, which can even embed a rigid-body alignment.}. For each subject, let $\vec{\phi}_n \in \mathbb{R}^{I \times 3}$ be the diffeomorphic transformation, and let \linebreak$\vec{\mu}_n = \mathrm{softmax}\left(\vec{a}\circ\vec{\phi}_n\right)$ be the deformed template. The likelihood of observed voxel values at locations $\left\{\vec{x}_i ~;~ 1 \leqslant i \leqslant I\right\}$ is:
\begin{equation}
    \textstyle
    \small
    p(\vec{f}_n(\vec{x}_i) \mid \vec{\mu}_n(\vec{x}_i)) = \mathrm{Cat}(\vec{f}_n(\vec{x}_i) \mid \vec{\mu}_n(\vec{x}_i)) = \prod_{k=1}^K \mu_n^{(k)}(\vec{x}_i)^{f_n^{(k)}(\vec{x}_i)} \enspace.
\end{equation}\par

In this work, diffeomorphisms are defined as geodesics, according to a Riemannian metric\footnote{In this work, it is a combination of membrane, bending and linear-elastic energies.} defined by a positive definite operator named $L$, and are thus entirely parameterised by their initial velocity \cite{Miller2006}. A complete transformation $\vec{\phi}$ is recovered by integrating the velocity in time, knowing that the momentum $\vec{u}_t = L \vec{v}_t$, is conserved at any $t$:
\begin{equation}
    \textstyle
    \small
    \vec{u}_t(\vec{x}) = 
    \left|\vec{D}\vec{\phi}_t^{-1}(\vec{x})\right| \left(\vec{D}\vec{\phi}_t^{-1}(\vec{x})\right)^{T}
    \left(\vec{u}_0\circ\vec{\phi}_t^{-1}(\vec{x})\right)
    \enspace.
\end{equation}%
The velocity field can be retrieved from the momentum field by performing the inverse operation $\vec{v}_t = K \vec{u}_t$, where $K$ is $L$'s Green's function Because we want to model inter-individual variability, we need them to be all defined in the same (template) space, which is achieved by using the initial velocity of their inverse\footnote{The initial velocity of $\vec{\phi}$ is the opposite of the final velocity of $\vec{\phi}^{-1}$, and \emph{vice versa}.}, that we name $\left\{\vec{v}_n \in \mathbb{R}^{I \times 3} ~;~ 1 \leqslant n \leqslant N\right\}$. Following \cite{Tipping1999b}, we use the probabilistic principal component analysis (PPCA) framework to regularise initial velocity fields, which leads to writing them as a linear combination of principal modes plus a residual field. Let us assume that we want to model them with $M \ll 3I$ principal components. Then, let $\vec{W}$ be a $3I \times M$ matrix (called the \emph{principal subspace}), each column being one principal component, let $\vec{z}_n$ be the latent representation of a given velocity field in the principal subspace and let $\vec{r}_n$ be the corresponding residual field. This yields $\vec{v}_n = \vec{W}\vec{z}_n + \vec{r}_n$. In \cite{Tipping1999b}, latent coordinates $\vec{z}_n$ stem from a standard Gaussian and $\vec{r}_n$ is i.i.d. Gaussian noise, and a maximum-likelihood estimate of the principal subspace is retrieved. A Bayesian version can be designed by placing a Gaussian prior on each principal component \cite{Bishop1999}. Smooth velocities can be enforced with a smooth prior over each principal component and over the residual field, and a Gaussian prior over the latent coordinates, yielding:
{\small\begin{align}
    p(\vec{W}) & = \prod_{m=1}^M \mathcal{N}\left(\vec{w}_m \mid \vec{0}, \vec{L}^{-1}\right) \enspace, \\
    p(\vec{z}_n \mid \vec{A}) & = \mathcal{N}\left(\vec{z}_n \mid \vec{0}, \vec{A}^{-1}\right) \enspace, \\
    p(\vec{r}_n \mid \lambda) & = \mathcal{N}\left(\vec{r}_n \mid \vec{0}, \left(\lambda\vec{L}\right)^{-1}\right) \enspace,
    \label{eq:reg1}
\end{align}}%
where $\vec{L}$ is the discretisation of $L$ and $\lambda$ is the anatomical noise precision. However, this approach is often not regularised enough. Zhang and Fletcher \cite{Zhang2015b} proposed a different prior, which can be seen as being set over the reconstructed velocities. In practice, it takes the form of a joint distribution over all latent coordinates, residual fields and the principal subspace:
\begin{equation}
    \textstyle
    \small
    p(\vec{z}_{1 \dots N}, \vec{r}_{1 \dots N}, \vec{W}) 
    \propto
    \prod_{n=1}^N \mathcal{N}\left(\vec{W}\vec{z}_n + \vec{r}_n \mid \vec{0}, \vec{L}^{-1}\right)
    \enspace.
    \label{eq:reg2}
\end{equation}%
The advantage of the first formulation (4-6) is that it explicates the covariance matrix of the latent coordinates and the noise precision, which can then be inferred from the data. Additionally, it could be extended to multimodal latent distributions such as Gaussian mixtures. However, the second formulation \eqref{eq:reg2} is better at effectively regularising the principal subspace and, in general, the reconstructed velocities. In practice, we use a weighted combination of these two formulations, and call the weights $\gamma_1$ and $\gamma_2$.

The noise precision, $\lambda$, can be inferred in a Bayesian fashion by introducing a conjugate Gamma prior\footnote{The Gamma prior is a parameterised such that $\mathbb{E}\left[\lambda\right] = \lambda_0$.} with $\alpha = \frac{\nu_0 \times 3I}{2}$ and $\beta = \frac{\nu_0 \times 3I}{2\lambda_0}$ as shown in \cite{Simpson2012}. Similarly, the latent covariance matrix is given a conjugate Wishart prior, which is made as non-informative as possible by setting its degrees of freedom to $M$, and whose expected value is the identity matrix, \emph{i.e.}, $p(\vec{A}) = \mathcal{W}\left(\vec{A} ~\middle|~ \frac{1}{M}\vec{I}, M\right)$. This prior has the opposite effect of an automatic relevance determination prior, since it prevents principal components from collapsing during the first iterations by promoting non-null variances.

\begin{figure}[t]
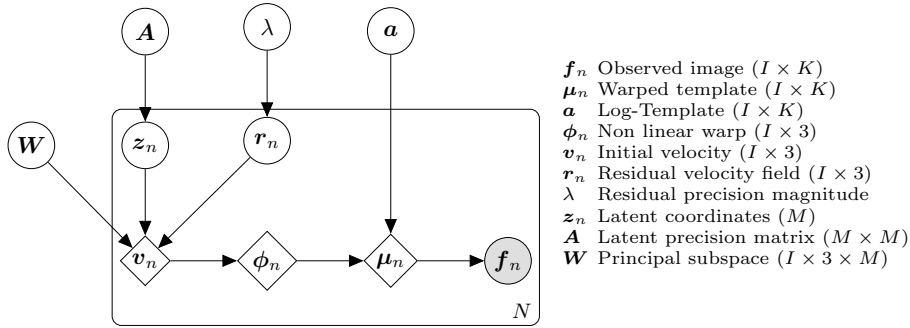

    \begin{minipage}[c]{0.58\textwidth}
    \centering
    \resizebox{\textwidth}{!}{%
    \tikz{%
        \node[obs] (f) {$\vec{f}_n$} ;%
        \node[det, left=of f] (mu) {$\vec{\mu}_n$} ;%
        \node[det, left=of mu] (phi) {$\vec{\phi}_n$} ;%
        \node[det, left=of phi] (v) {$\vec{v}_n$} ;%
        \node[latent, above=of phi] (r) {$\vec{r}_n$} ;%
        \node[latent, above=of r] (l) {$\lambda$} ;%
        \node[latent, above=of v] (z) {$\vec{z}_n$} ;%
        \node[latent, left=of z] (W) {$\vec{W}$} ;%
        \node[latent, above=of z] (Az) {$\vec{A}$} ;%
        \node[latent, above=of mu, yshift=1.7cm] (a) {$\vec{a}$} ;%
        \edge {a, phi} {mu} ;%
        \edge {mu} {f} ;%
        \edge {v} {phi} ;%
        \edge {W, z, r} {v} ;%
        \edge {l}{r} ;%
        \edge {Az} {z} ;%
        \plate {N} {(f)(mu)(phi)(v)(z)(r)} {$N$} ;%
    }}
    \end{minipage}
    \hspace{3pt}
    \begin{minipage}[c]{0.42\textwidth}
    \scriptsize
        \begin{tabular}{ll}
            $\vec{f}_n$ & Observed image ($I \times K$) \\
            $\vec{\mu}_n$ & Warped template ($I \times K$) \\
            $\vec{a}$ & Log-Template ($I \times K$) \\
            $\vec{\phi}_n$ & Non linear warp ($I \times 3$) \\
            $\vec{v}_n$ & Initial velocity ($I \times 3$) \\
            $\vec{r}_n$ & Residual velocity field ($I \times 3$) \\
            $\lambda$ & Residual precision magnitude \\
            $\vec{z}_n$ & Latent coordinates ($M$) \\
            $\vec{A}$ & Latent precision matrix ($M \times M$) \\
            $\vec{W}$ & Principal subspace ($I \times 3 \times M$)
        \end{tabular}
    \end{minipage}
    \caption{Generative shape model, in the form of a graphical model. Circles indicate random variables while diamonds indicate deterministic variables. Shaded variables are observed. Plates indicate replication.}
    \label{fig:bnet}
\end{figure}

Finally, we look for a maximum \emph{a posteriori} estimate of the template, $\vec{a}$, with a very uninformative log-Dirichlet prior that prevents null probabilities (a smooth prior could also be used following \cite{Ashburner2009}). The complete model is depicted in the form of a graphical model in Fig. \ref{fig:bnet}.

\subsection[Inference]{Inference\protect\footnote{$q$ is used for approximate posteriors and $\mathbb{E}_q$ for posterior expected values. Superscript stars denote optimal approximations. $\cequal$ means ``equal up to an additive constant''.}}

A basic inference scheme would be to search for a mode of the model likelihood, by optimising in turn each parameter of the model. It is however more consistent to tackle this problem as one of missing data, which is dealt with by computing the posterior distribution over all latent variables. Unfortunately, the posterior does not possess a tractable form. A solution is to use variational inference to describe an approximate posterior $q$ that can be more easily computed, by restricting the search space to distributions that factorise over a subset of variables \cite{Bishop2006}. This method allows the uncertainty about parameters estimates to be accounted for when inferring other parameters. Here, for computational reasons, we do not perform a fully Bayesian treatment of the problem and look for mode estimates of the principal subspace and template. We still marginalise over all subject-specific parameters (latent coordinates and residual field), as recommended by \cite{Allassonniere2007a}. We state that the set of marginalised latent variables is \linebreak$\vec{\Upsilon} = \left\{\vec{z}_{1 \dots N}, \vec{r}_{1 \dots N}, \vec{A}, \lambda\right\}$ and make the (mean field) approximation that the posterior factorises over $(\vec{z}_{1 \dots N})$, $(\vec{r}_{1 \dots N})$, $(\vec{A})$ and  $(\lambda)$.

Since we used conjugate priors for the latent precision matrix and the anatomical noise precision, their posterior have the same form as their prior and update equations are equivalent to computing a weighted average between their prior expected value and their maximum likelihood estimator. In contrast, posterior distributions of $\vec{z}_n$ and $\vec{r}_n$ have no simple form. We thus make a Laplace approximation and estimate their mean ($\vec{z}_n^\star$, $\vec{r}_n^\star$) and covariance ($\vec{S}_{\mathrm{z},n}^\star$, $\vec{S}_{\mathrm{r},n}^\star$) with their mode and second derivatives about this mode. They are obtained by Gauss-Newton optimisation and the corresponding derivations are provided in Sec. \ref{sec:gn}. Because of the non-linearity induced by geodesic shooting and template interpolation, we first make the approximation that:
{\small\begin{align}
    \mathbb{E}_q\Big[p\left(\vec{f}_n \mid \vec{z}_n, \vec{r}_n, \vec{W}, \vec{a}\right)\Big] 
    \approx p\left(\vec{f}_n \mid \vec{z}_n^\star, \vec{r}_n^\star, \vec{W}, \vec{a}\right)
    = p\left(\vec{f}_n \mid \vec{\mu}_n^\star\right)
    \enspace,
\end{align}}%
where $\mathbb{E}_q$ means the posterior expected value and $\vec{\mu}_n^\star$ is the template deformed according to the above parameter estimates. Consequently, we find:
{\small\begin{align}
    \ln q^\star(\vec{z}_n) 
    & \cequal \ln p\left(\vec{f}_n \mid \vec{\mu}_n^\star \right) - \frac{1}{2}\vec{z}_n^{T}\left(\gamma_1 \vec{A} + \gamma_2 \vec{W}^{T}\vec{L}\vec{W}\right)\vec{z}_n - \gamma_2\vec{z}_n^{T}\vec{W}^{T}\vec{L}\vec{r}_n^\star
    \\
    \ln q^\star(\vec{r}_n)
    & \cequal \ln p\left(\vec{f}_n \mid \vec{\mu}_n^\star \right) - \frac{\gamma_1\lambda^\star + \gamma_2}{2}\vec{r}_n^{T}\vec{L}\vec{r}_n - \gamma_2\vec{r}_n^{T}\vec{L}\vec{W}\vec{z}_n^\star
    \enspace.
\end{align}}\par

\subsection{Gauss-Newton optimisation}
\label{sec:gn}

Gauss-Newton (GN) optimisation of an objective function $\mathcal{E}$ with respect to a vector of parameters $\vec{x}$ consists of iteratively improving the objective function by making, locally, a  second-order approximation. The gradient, $\vec{g}$, and Hessian matrix, $\vec{H}$, are computed about the current best estimate of the optimal parameters, $\vec{x}_i$, and the new optimum is found according to $\vec{x}_{i+1} = \vec{x}_i - \vec{H}^{-1} \vec{g}$. In practice, this update scheme sometimes overshoots it is therefore common to perform a backtracking line search along the direction $-\vec{H}^{-1} \vec{g}$.

\subsubsection*{Differentiating the data term:} Let us write $\mathcal{E}$ the (negative) data term for an arbitrary subject:
\begin{equation}
    \textstyle
    \small
    \mathcal{E}
    = -\ln p\left(\vec{f} \mid \vec{\mu}\right) 
    = -\sum_{i=1}^{I} \ln\mathrm{Cat}\Big(\vec{f}(\vec{x}_i) ~\Big|~ \mathrm{softmax}\left(\vec{\Phi}\vec{a}(\vec{x}_i)\right)\Big)
    = \mathcal{C}_{\vec{f}}\left(\vec{\Phi}\vec{a}\right)
    \enspace.
\end{equation}
Following \cite{Ashburner2011a}, differentiating $\mathcal{E}$ with respect to $\vec{v}$ can be approximated by differentiating with respect to $\vec{\Phi}$ and applying the chain rule, which yields:
\begin{equation}
    \textstyle
    \small
    \frac{\partial \mathcal{E}}{\partial \vec{v}} = \left(\vec{\Phi}^{T} \vec{\nabla} \mathcal{C}_{\vec{f}}\left(\vec{\Phi}\vec{a}\right)\right) \vec{\nabla}\vec{a} \enspace,
\end{equation}
where $\vec{\nabla} \mathcal{C}_{\vec{f}}$ is the gradient of the log-Categorical distribution and takes the form of an $I \times K$ vector field, and $\vec{\nabla}\vec{a}$ contains spatial gradients of the log-template and takes the form of an $I \times 3$ vector field. Second derivatives can be approximated by:
\begin{equation}
    \textstyle
    \small
    \frac{\partial^2 \mathcal{E}}{\partial \vec{v}^2} = \vec{\nabla}\vec{a}^{T}\left(\vec{\Phi}^{T} \vec{\nabla}^2 \mathcal{C}_{\vec{f}}\left(\vec{\Phi}\vec{a}\right)\right) \vec{\nabla}\vec{a} \enspace,
\end{equation}
where $\vec{\nabla}^2 \mathcal{C}_{\vec{f}}$ is the Hessian of the log-Categorical distribution and takes the form of an $I \times K \times K$ symmetric tensor field. The gradient and Hessian of $\mathcal{C}_{\Vec{f}}$ were derived in \cite{Ashburner2009} and can be computed in each voxel according to:
\begin{equation}
    \textstyle
    \small
    \frac{\partial \mathcal{C}_{\vec{f}}(\vec{a})}{\partial a_k} 
    = \mu_k \left(\sum_{l=1}^K f_l\right) - f_k \enspace, \enspace
    \frac{\partial^2 \mathcal{C}_{\vec{f}}(\vec{a})}{\partial a_k \partial a_m}
    = \mu_k(\delta^m_k - \mu_m) \sum_{l=1}^K f_l \enspace.
\end{equation}
Since $\vec{v} = \vec{W}\vec{z} + \vec{r}$, derivatives with respect to $\vec{r}$, $\vec{z}$ and $\vec{W}$ are obtained by applying the chain rule. 

\subsubsection*{Orthogonalisation:} The PPCA formulation is invariant to rotations inside the latent space \cite{Tipping1999b}. Consequently, it allows finding an optimal subspace but does not enforce the individual bases $\vec{w}_{1 \dots M}$ to be the eigenmodes of the complete covariance matrix. It makes sense, however, to transform the subspace so that it corresponds to the first eigenmodes as it eases the interpretation. Also, in order to enforce a sparse Hessian matrix over $\vec{W}$, we require $\vec{Z}\vec{Z}^{T}$ to be diagonal, where the columns of $\vec{Z}$ are the individual $\vec{z}_n$. This leads us to look for an $M \times M$ transformation matrix $\vec{T}$ that keeps the actual diffeomorphisms untouched ($\vec{W}\vec{z} = \vec{W}\vec{T}^{-1}\vec{T}\vec{z}$), while diagonalising both $\vec{Z}\vec{Z}^{T}$ and $\vec{W}^{T}\vec{L}\vec{W}$. This is done by a series of singular value decompositions that insure that $\vec{T}\vec{Z}\vec{Z}^{T}\vec{T}^{T}$ is diagonal and $\vec{T}^{-T}\vec{W}^{T}\vec{L}\vec{W}\vec{T}^{-1}$ is the identity. However, the distribution of diagonal weights between $\vec{W}^{T}\vec{L}\vec{W}$ and $\vec{Z}\vec{Z}^{T}$ is not optimal and we optimise an additional diagonal scaling matrix $\vec{Q}$ by alternating between updating $\vec{A}$ from the rotated $\mathbb{E}\left[\vec{Z}\vec{Z}^{T}\right]$, and updating the scaling weights by Gauss-Newton optimisation of the remaining terms of the lower bound that depend on them:
\begin{equation}
    \small
    \textstyle
    \mathcal{E}_{\Vec{Q}} = -\frac{1}{2}\Big(\mathrm{Tr}\left(\vec{Q}\vec{T}\vec{Z}\vec{Z}^{T}\vec{T}^{T}\vec{Q}\vec{A}\right) + \mathrm{Tr}\left(\vec{Q}^{-1}\vec{T}^{-T}\vec{W}^{T}\vec{L}\vec{W}\vec{T}^{-1}\vec{Q}^{-1}\right)\Big) \enspace.
\end{equation}

\section{Experiments and results}

We ran the algorithm on a training set consisting of the first 38 subjects of the OASIS cross-sectional database \cite{Marcus2007}. We used the provided FSL segmentations, which we transformed into tissue probability maps by extracting the grey and white matter classes and smoothing them with a 1-voxel FWHM Gaussian kernel. We set the number of principal components to 32, the parameters of the membrane, bending and linear-elastic energies were respectively set to 0.001, 0.02 and $(0.0025, 0.005)$, and we used $\gamma_1 = \gamma_2 = 1$. We set the prior parameters of the residual precision magnitude based on tests conducted on 2D axial slices ($\lambda_0 = 17$, $\nu_0 = 10$). Templates reconstructed with a varying number of principal modes, and with or without the residual field, are presented in Fig. 2, while Fig. 3 shows the template deformed along the first two principal modes, the first one being typical of brain ageing. This pattern is validated by plotting coordinates along the first dimension against actual ages. Finally, the learnt model was tested by registering the template towards 38 unseen images from the OASIS database. The distribution of categorical log-likelihood values for the training and testing sets are depicted in Fig. 4, along with two example fits that were randomly selected. Both sets have similar distributions (mean $\pm$ std $\times 10^5$. Train: $-7.23 \pm 1.26$ ; Test: $-7.55 \pm 0.88$), showing the model's robustness.

\begin{figure}[!h]
    \centering
    \includegraphics[width=\textwidth]{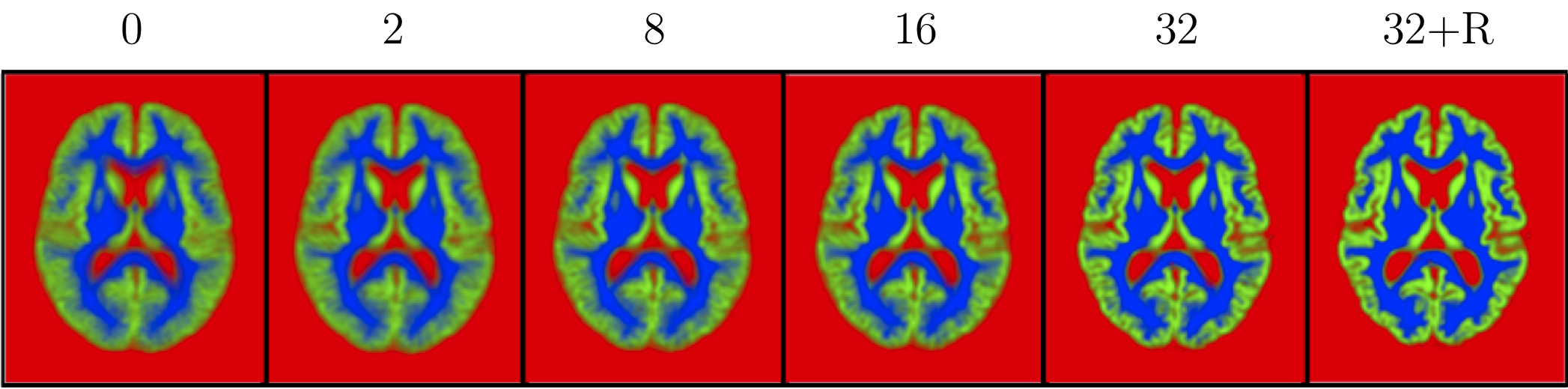}
    \caption{Template reconstructed using an increasing number of principal modes, and with the addition of the residual field.}
    \includegraphics[width=\textwidth]{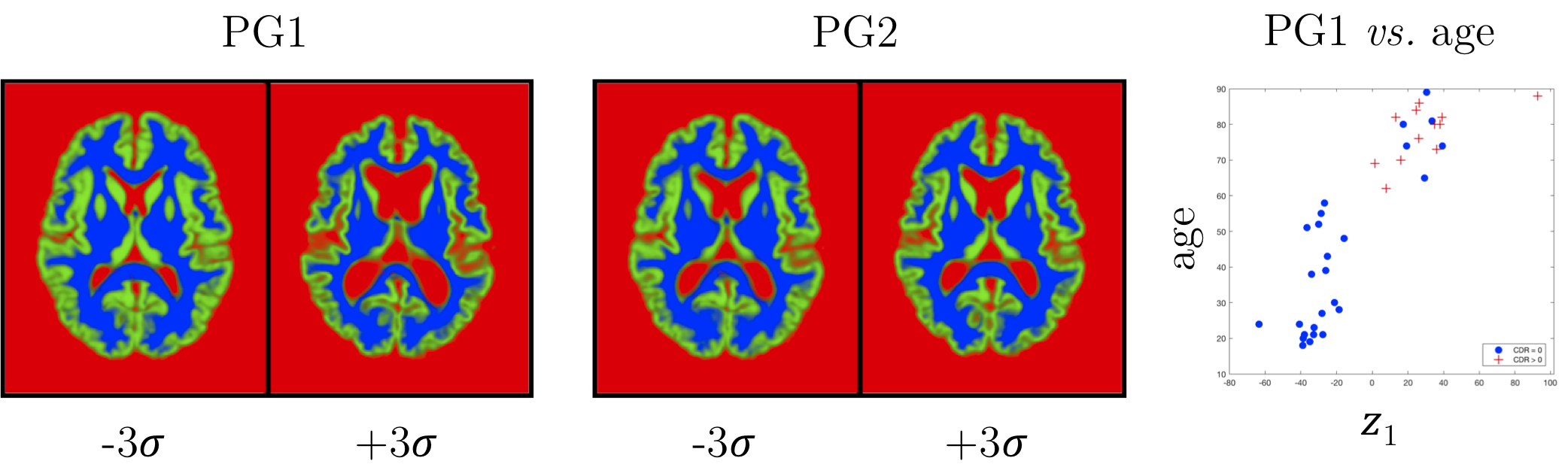}
    \caption{Left: deformed template along the first two principal modes. Right: latent coordinates \emph{vs.} subject ages. Probable AD subjects are marked with a red cross.}
    \includegraphics[width=\textwidth]{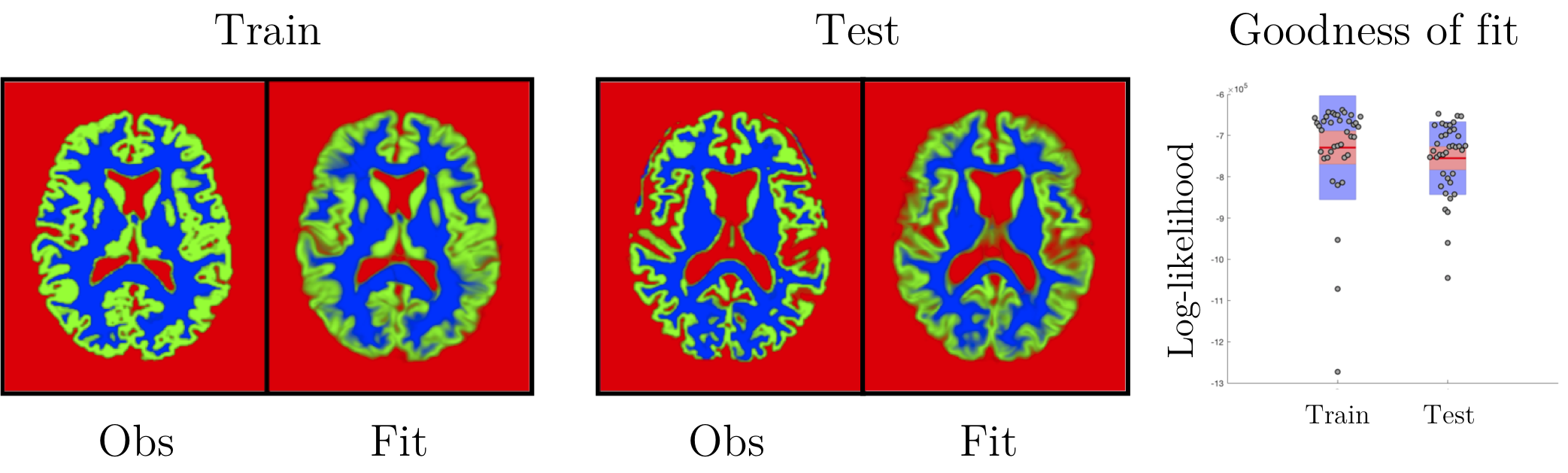}
    \caption{Left: two random examples of fit from the train and test sets. Right: distribution of categorical log-likelihood values for the train and test sets.}
\end{figure}


\section{Conclusion}

We presented a generative model of brain shape that does not limit the space of diffeomorphisms, allowing learning regularisation while preserving enough flexibility for accurate normalisation. We showed how principal modes of variation correlate with known factors of brain shape variability, hinting towards the fact that this low-dimensional representation might allow to discriminate between physiological states. Future research will focus on applying this framework to very large databases, and on combining it with segmentation models in order to work directly with raw data. The latent distribution may be improved by using multimodal priors such as Gaussian mixtures. Our main limitation is the small number of principal basis that can be learned due to their large size. This could be overcome by explicitly modelling sparse and local covariance patterns.

\bibliographystyle{splncs03}
\bibliography{miccai2018}

\end{document}